\documentclass[10pt,twocolumn,letterpaper]{article}

\usepackage{cvpr}
\usepackage{times}
\usepackage{epsfig}
\usepackage{graphicx}
\usepackage{amsmath}
\usepackage{amssymb}
\usepackage{bm}
\usepackage{esvect}
\usepackage{amsbsy}
\usepackage{multirow}
\usepackage[export]{adjustbox}
\usepackage{hhline}
\usepackage{makecell}

\usepackage[pagebackref=true,breaklinks=true,letterpaper=true,colorlinks,bookmarks=false]{hyperref}

\usepackage{array,booktabs}
\newcolumntype{C}[1]{>{\centering\let\newline\\\arraybackslash\hspace{0pt}}p{#1}}
\cvprfinalcopy 

\def\cvprPaperID{3292} 

\ifcvprfinal\fi
\begin{document}


\title{MX-LSTM: mixing tracklets and vislets to jointly forecast\\ trajectories and head poses}

\author{Irtiza Hasan$^{1,2}$, Francesco Setti$^{1}$, Theodore Tsesmelis$^{1,2,3}$, Alessio Del Bue$^{3}$,\\ Fabio Galasso$^{2}$, and Marco Cristani$^{1}$ \vspace{.1em}\\
{\normalsize $^{1}$ University of Verona (UNIVR) \hspace{.05\textwidth} $^{2}$ OSRAM GmbH  \hspace{.05\textwidth} $^{3}$ Istituto Italiano di Tecnologia (IIT)}\\
{\tt\small \{irtiza.hasan,francesco.setti,marco.cristani\}@univr.it}, \\ 
{\tt\small \{t.tsesmelis,f.galasso\}@osram.com, alessio.delbue@iit.it}
}

\maketitle
\thispagestyle{empty}


\newcommand{\todo}[1]{}
\renewcommand{\todo}[1]{{\color{red} TODO: {#1}}}
\newcommand{\bo}[1]{\boldsymbol{#1}}

\newcommand{\fs}[1]{{\color{red} [FRANZ: {#1}]}}


\begin{abstract} \label{sec:abstract} 
    Recent approaches on trajectory forecasting use tracklets to predict the future positions of pedestrians exploiting Long Short Term Memory (LSTM) architectures.
    This paper shows that adding \emph{vislets}, that is, short sequences of head pose estimations, allows to increase significantly the trajectory forecasting performance.
    We then propose to use vislets in a novel framework called MX-LSTM,  
    capturing the interplay between tracklets and vislets thanks to a joint unconstrained optimization of full covariance matrices during the LSTM backpropagation.
    At the same time, MX-LSTM predicts the future head poses, increasing the standard capabilities of the long-term trajectory forecasting approaches.
    With standard head pose estimators and an attentional-based social pooling, MX-LSTM scores the new trajectory forecasting state-of-the-art in all the considered datasets (Zara01, Zara02, UCY, and TownCentre) 
    with a dramatic margin when the pedestrians slow down, a case where  most of the forecasting approaches struggle to provide an accurate solution.
\end{abstract}

\vspace{-0.1cm}
\section{Introduction} \label{sec:intro}
\vspace{-0.1cm}
Anticipating the trajectories that could occur in the future is important for several reasons: in computer vision, path forecasting helps the dynamics modeling for target tracking~\cite{pellegrini2009iccv,robicquet2016eccv,sadeghian2017tracking,yamaguchi2011cvpr} and behavior understanding~\cite{alahi2016cvpr,kitani2012activity,lee2016predicting,ma2017forecasting,robicquet2016eccv}; in robotics, autonomous systems should plan routes that will avoid collisions and be respectful of the human proxemics~\cite{dragan2011manipulation,hall1966hidden,kuderer2012feature,mainprice2016goal,trautman2010iros,ziebart2009planning}. Recently, path forecasting has benefited from the introduction of Long Short Term Memory (LSTM) architectures~\cite{alahi2016cvpr,ijcai2017-386,hochreiter1997long,su2016crowd,sun20173dof,varshneya2017human}.

All of these approaches use exclusively the $(x,y)$ position coordinates for the prediction, forgetting that humans act and react using their senses to explore the environment, in particular, through the visual information conveyed by the gaze and inferred by the head pose~\cite{caminada1980philips,chen2012we,davoudian2012pedestrians,fotios2015using,fotios2015usingII,foulsham2011and,intriligator2001spatial,patla2003far,robertson2006estimating,stiefelhagen1999gaze,vansteenkiste2013visual}. 
In particular, ~\cite{caminada1980philips,davoudian2012pedestrians,fotios2015using,fotios2015usingII,foulsham2011and,intriligator2001spatial,patla2003far,vansteenkiste2013visual} found that the head pose correlates to the person destination and pathway: these findings are also supported by a statistical analysis presented in our paper (Sec. \ref{sec:dataset}).


For the first time this work considers the head pose, jointly with the positional information, as a cue to perform forecasting. In particular, tracklets (sequences of $(x,y)$ coordinates) and \emph{vislets}, that is, reference points indicating the head pan orientation, are the input of the novel 
MiXing LSTM (MX-LSTM), an LSTM-based model that learns how tracklet and vislet streams are related, mixing them together in the LSTM hidden state recursion by means of cross-stream full covariance matrices, optimized during backpropagation.


MX-LSTM is able to encode how movements of the head and the people dynamics are connected. For example, it captures the fact that rotating the head towards a particular direction may anticipate a trajectory drifting with an acceleration (as in the case of a person leaving a group after a conversation).
This happens thanks to a novel optimization of the LSTM parameters using a Gaussian full covariance through an unconstrained log-Cholesky parameterization in the backpropagation, securing  positive semidefinite  matrices. To the best of our knowledge, this is the first time Gaussian distributions with covariance matrices of order higher than two are optimized in LSTMs.

Vislet information is also used to build a scene context, i.e. where are the other people and how they are moving, by a shared state pooling as in \cite{alahi2016cvpr,varshneya2017human}, that here is further improved using the head pose by discarding the people that an individual cannot see.

As a by-product, MX-LSTM predicts head orientations too, allowing to reason where  people will most probably look at, providing a fine grained level of long-term prediction never reached so far in crowded scenarios.

Adopting standard protocols for trajectory forecasting
~\cite{alahi2016cvpr,lerner2007crowds,pellegrini2009iccv} and using head poses information given by a standard head pose estimator~\cite{lee2015fast}, MX-LSTM defines the new state-of-the-art both in the UCY sequences (Zara01, Zara02 and UCY) and in the TownCentre dataset. In particular, MX-LSTM has the ability to forecast people when they are moving slowly, the Achille's heel of all the other approaches proposed so far. 

As main contributions, in this paper:
\begin{itemize}
    \item \vspace{-5pt} We show that trajectory forecasting can be dramatically ameliorated by considering head pose estimates;\vspace{-5pt}
    \item We propose a novel LSTM architecture, MX-LSTM, which exploits positional (tracklets) and orientational (vislets) information thanks to an optimization of $d$-variate Gaussian parameters including full covariances with $d>2$;\vspace{-5pt}
    \item We motivate the need for MX-LSTM showing that head poses are related with the trajectories, even at low velocities, where most of the forecasting approaches fail;\vspace{-5pt}
    \item We define a novel type of social pooling, in the sense of~\cite{alahi2016cvpr,varshneya2017human}, by exploiting the vislet information;\vspace{-5pt}
    \item Thanks to MX-LSTM, we define state-of-the-art forecasting results on different datasets;\vspace{-5pt}
    \item We present MX-LSTM results of head pose forecasting, showing new long-term behavior analysis capabilities.\vspace{-5pt}
\end{itemize}

The rest of the paper is organized as follows.
Sec. \ref{sec:prev} reviews the related literature. Sec.~\ref{sec:our} presents the proposed MX-LSTM while Sec.~\ref{sec:dataset} motivates its design by showing how head pose and trajectories are related in the most po\-pu\-lar forecasting datasets. We show quantitative and qualitative experiments in Sec.~\ref{sec:experiments}, concluding the paper in Sec.~\ref{sec:conc}.

\section{Related work} \label{sec:prev}
\vspace{-0.1cm}

Classical forecasting approaches~\cite{morris2008survey} adopted Kalman filters~\cite{kalman1960new}, linear~\cite{mccullagh1989generalized} or Gaussian regression models~\cite{quinonero2005unifying,rasmussen2006gaussian,wang2008gaussian,williams1998prediction}, autoregressive models~\cite{akaike1969fitting} and time-series analysis~\cite{priestley1981spectral}. These approaches ignore  human-human interactions, which instead play a major role in  recent literature. 

\noindent\textbf{Human-human interactions.}
The consideration of other pedestrians in the scene and their innate avoidance of collisions was first pioneered by~\cite{helbing1995social}.
This initial seed was further developed by~\cite{lerner2007crowds}, \cite{ma2017forecasting} and \cite{pellegrini2009iccv}, which respectively introduced a data-driven, a continuous, and a game theoretical model.
Notably, these approaches successfully employ essential cues for track prediction such as the human-human interaction and the people intended destination.
More recent works encode the human-human interactions into a ``social'' descriptor~\cite{alahi2014socially} or propose human attributes~\cite{yi2015understanding} for the forecasting in crowds.
More implicitly, other methods~\cite{alahi2016cvpr,varshneya2017human} embed proxemic reasoning in the prediction by pooling hidden variables representing the probable location of a pedestrian in a LSTM.
Our work mainly differentiates from~\cite{alahi2016cvpr,lerner2007crowds,pellegrini2009iccv,varshneya2017human} because we only consider for interactions those people who are within the cone of attention of the person, (as also verified by psychological studies~\cite{intriligator2001spatial}). 

\noindent\textbf{Destination-focused path forecast.}
%
Path forecasting has also been framed as an inverse optimal control (IOC) problem~\cite{kitani2012activity}. Follow-up work adopted inverse reinforcement learning~\cite{abbeel2004apprenticeship,ziebart2008maximum} and dynamic reward functions~\cite{lee2016predicting} to address the occurring changes in the environment. We describe these approaches as destination-focused because they all require the end-point of the person track to be known, which later works have relaxed to a set of plausible path ends~\cite{dragan2011manipulation,mainprice2016goal}. Here we discard this information, that in our opinion undermines the reason why we may be predicting the tracks.

\noindent\textbf{The head pose and the social motivation.}
The interest into the head pose stems from sociological studies such as \cite{caminada1980philips,davoudian2012pedestrians,fotios2015using,fotios2015usingII,foulsham2011and,patla2003far,vansteenkiste2013visual}, whereby head pose has been shown to correlate to the person destination and pathway. 
In this paper, we also discover that the head pose is correlated with the movement, especially at high velocities, while slowing down this correlation decreases too, but still remaining statistically significant. These studies motivate the use of the head pose as a proxy to the  track forecasting.\\
Using head pose comes with the further advantage that it can be estimated at small resolutions~\cite{ba2004probabilistic,gourier2006head,hasan2017tiny,lee2015fast,robertson2006estimating,stiefelhagen1999gaze,tosato2013characterizing}, thus requiring no oracle information and enabling a real-time system. Without loss of generality, for the head pose estimation we adopt the publicly available algorithm of~\cite{lee2015fast}.

\noindent\textbf{LSTM models.}
LSTM models~\cite{hochreiter1997long} are employed in those tasks where the output is conditioned on a varying number of inputs~\cite{gregor2015draw,vinyals2015show}, notably hand writing generation~\cite{graves2013generating} and tracking~\cite{coskun2017long}.

%
%
%

As for trajectory forecasting,~\cite{alahi2016cvpr} models pedestrians as LSTMs that share their hidden states through a ``social'' pooling layer, avoiding to forecast colliding trajectories. This idea has been successfully adopted by~\cite{varshneya2017human}, and further developed in \cite{sadeghian2017tracking} for modeling the tracking dynamics. A similar idea has been embedded directly in the LSTM memory unit as a regularization that models the local spatial and temporal dependency between neighboring pedestrians~\cite{ijcai2017-386,su2016crowd}. 
As written above, here we modify the social pooling by considering a visibility attentional area driven by the head pose. 


In most of the cases, the training of LSTMs for forecasting minimizes the negative log-likelihood over Gaussians~\cite{alahi2016cvpr,varshneya2017human} or mixture of Guassians~\cite{graves2013generating}. In general, when it comes to Gaussian log-likelihood loss functions, only bidimensional data (\ie $(x,y)$ coordinates) have been considered so far, leading to the estimation of $2\times2$ covariance matrices. These can be optimized without considering the positive semidefinite requirement~\cite{graves2012supervised}, which is one of the most important problems for the covariances obtained by optimization~\cite{pinheiro1996unconstrained} (see Sec.~\ref{sec:opt}). Here for the first time, we study the problem of optimizing Gaussian parameters of higher dimensionality. 


\section{Our approach}\label{sec:our}
\vspace{-0.1cm}

In this section we present the \emph{MX-LSTM}, capable of jointly forecasting positions and head orientations of an individual thanks to the presence of two information streams: Tracklets and \emph{vislets}.

\subsection{Tracklets and vislets}
\begin{figure}[!ht] 
	\begin{center}
		\includegraphics[width=1\linewidth]{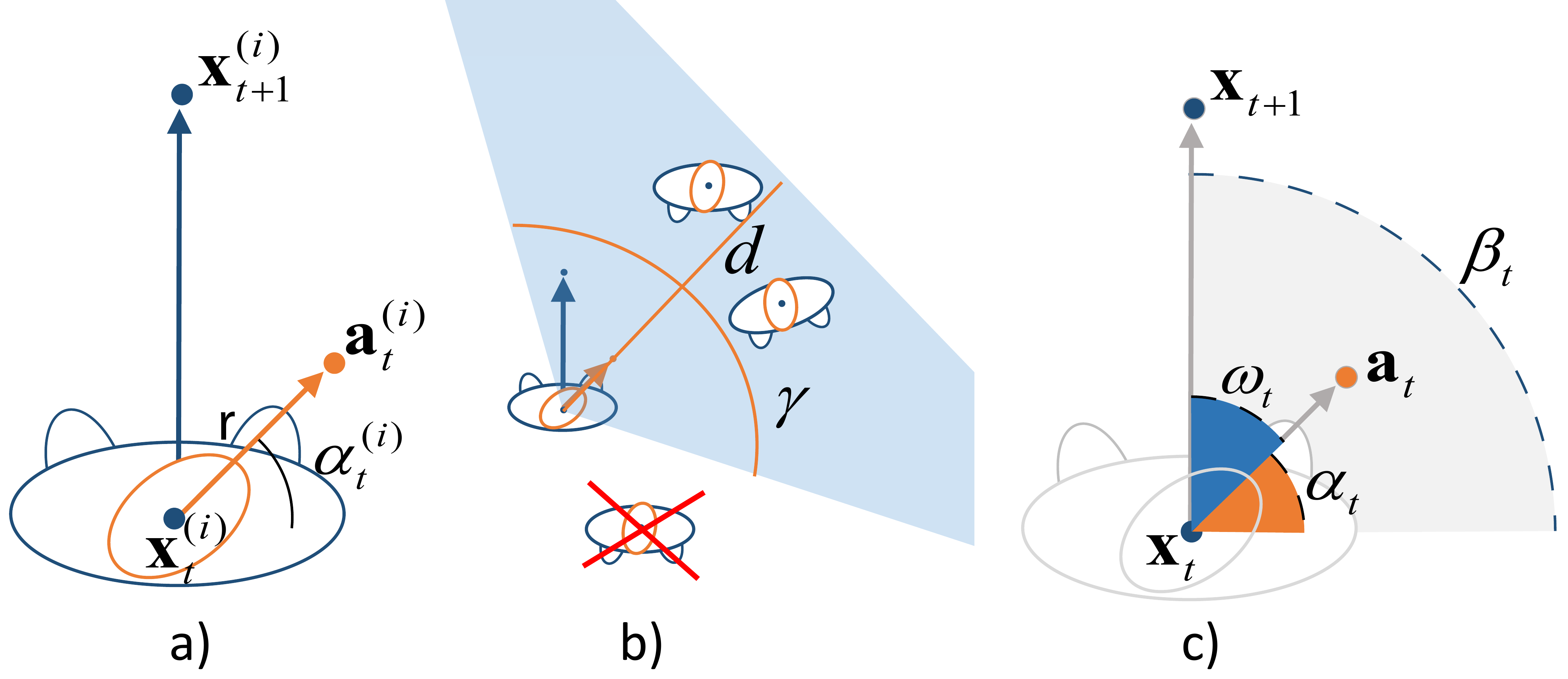}
	\caption{Graphical explanations; a) tracklets $\mathbf{x}^{(i)}_t$ and $\mathbf{x}^{(i)}_{t+1}$ and vislet anchor point $\mathbf{a}^{(i)}_t$; b) visual frustum pooling; c) angles for the correlation analysis (see Sec.~\ref{sec:dataset}).}\label{fig:explanations}
	\end{center}
    \end{figure}

\vspace{-0.1cm}
Given a subject $i$, a tracklet (see Fig.~\ref{fig:explanations}a)~) is formed by consecutive $(x,y)$ positions on the ground plane, $\{\mathbf{x}^{(i)}_t\}_{t=1,...,T}$, $\mathbf{x}^{(i)}_t = (x,y)\in \mathcal{R}^2$, while a vislet is formed by anchor points $\{\mathbf{a}^{(i)}_t\}_{t=1,...,T}$, with $\mathbf{a}^{(i)}_t = (a_x,a_y) \in \mathcal{R}^2$ indicating a reference point at a fixed distance $r$ from the corresponding $\mathbf{x}^{(i)}_t$, towards which the face is oriented\footnote{The distance $r$ is not influent in this work, and it can be any value; in this work we set it at $0.5$ for the visualization sake.}. 
In practice, $\mathbf{a}^{(i)}_t$ is a fixed size vector originating from $\mathbf{x}^{(i)}_t$, whose direction implicitly indicates the pan angle $\alpha^{(i)}_t$ of the head. In principle, it would be possible to encode the head orientation directly with an angle. We prefer the vislet representation because it does not show discontinuity (between 360$^\circ$ and 0$^\circ$) and because it is closer to the $(x,y)$ position representation and therefore more convenient for the vislet-position interplay. 

In the forecasting literature~\cite{alahi2016cvpr,trautman2010iros,yamaguchi2011cvpr} it is assumed that the prediction follows an ``observation'' period in which ground-truth data is fed into the machine. Here, the observation tracklets and vislets are fed into the MX-LSTM, which mixes together the two streams to understand their relationship, providing a joint prediction. In the experiments we evaluate the cases in which the past vislets are ground truth, but also the ``in-the-wild'' case, in which head pose is given by a real head detector. In this way, MX-LSTM will require no additional annotations in respect with former approaches.

A single MX-LSTM is instantiated for each pedestrian $i$, accepting tracklets and vislets with two separate embedding functions:
\begin{eqnarray}
    \mathbf{e}_{t}^{(x,i)} &=& \phi \left( \mathbf{x}^{(i)}_t,\mathbf{W}_{x} \right) \label{eq:embxy}\\
    \mathbf{e}_{t}^{(a,i)} &=& \phi \left( \mathbf{a}^{(i)}_t,\mathbf{W}_{a} \right) \label{eq:embvis}
\end{eqnarray}
where the embedding function $\phi$ consists in a linear projection through the embedding weigths $\mathbf{W}_{x}$ and $\mathbf{W}_{a}$ into a $D$-dimensional vector, multiplied by a RELU nonlinearity, where $D$ is the dimension of the hidden space.

\subsection{VFOA social pooling}
\vspace{-0.1cm}
The social pooling introduced in~\cite{alahi2016cvpr} is an effective way to let the LSTM capture how people move in a crowded scene avoiding collisions.  This work considers an isotropic interest area around the single pedestrian, in which the hidden states of the the neighbors are considered, including those which are \emph{behind} the pedestrian. In our case,  we improve this module using the vislet information by selecting which individuals to consider, by building a view frustum of attention (VFOA), that is a triangle originating from $\mathbf{x}^{(i)}_t$, aligned with $\mathbf{a}^{(i)}_t$, and with an aperture given by the angle $\gamma$ and a depth $d$; these parameters have been learned by cross-validation on the training partition of the TownCentre dataset (see Sec.~\ref{sec:experiments}).


Our view-frustum social pooling is a $N_o\times N_o \times D$  tensor, in which the space around the pedestrian is divided into a grid of $N_o\times N_o$
cells as in~\cite{alahi2016cvpr}, in which the VFOA is located, acting as the new interest region where people have to be taken into account. The pooling occurs as follows: 
\begin{equation}
    \mathbf{H}^{(i)}_t(m,n,:)=\sum_{j\in {VFOA}_i}\mathbf{h}^{(j)}_t,\label{eq:social}
\end{equation}
where the $m$ and $n$ indices run over the $N_o\times N_o$ grid and the condition $j\in {VFOA}_i$ is satisfied when the subject $j$ is in the VFOA of subject $i$. 
The pooling vector is then embedded into a $D$-dimensional vector by 
\begin{equation}
    \mathbf{e}_{t}^{(H,i)} = \phi({\mathbf{H}^{(i)}_t,\mathbf{W}_H}).\label{eq:pooling}
\end{equation}
Finally, the MX-LSTM recursion equation is
\begin{equation}
    \mathbf{h}^{(i)}_t=LSTM\left(\mathbf{h}^{(i)}_{t-1},\mathbf{e}_{t}^{(x,i)},\mathbf{e}_{t}^{(a,i)},\mathbf{e}_{t}^{(H,i)},\mathbf{W}_{\text{LSTM}}\right).
\end{equation}

\subsection{LSTM recursion}
\vspace{-0.1cm}
In principle (but in the next subsection we will ultimately modify the formulation), the hidden state is enforced to contain the parameters of a four dimensional Gaussian multivariate distribution $\mathcal{N}(\mathbf{\mu}^{(i)}_t,\mathbf{\Sigma}^{(i)}_t)$ as follows:
\begin{equation}
    [\mathbf{\mu}^{(i)}_t,\hat{\mathbf{\Sigma}}^{(i)}_t] = \mathbf{W}_o \mathbf{h}^{(i)}_{t-1},
\end{equation}
where $\hat{\mathbf{\Sigma}}^{(i)}_t$ is the vectorized version of $\mathbf{\Sigma}^{(i)}_t$.
In practice $\mathbf{\mu}^{(i)}_t = [\mu^{(x,i)}_t,\mu^{(y,i)}_t,\mu^{(a_x,i)}_t,\mu^{(a_y,i)}_t]$ and $\mathbf{\Sigma}^{(i)}_t$ contains the covariances among the $(x,y)$ coordinate distributions of the tracklets and the vislets.
The distribution is then sampled to generate the joint prediction of tracklets and vislet points $[\mathbf{\hat x}_t,\mathbf{\hat a}_t]$. In other words, we are able at the same time of forecasting trajectries and head poses.

The weight parameters of the LSTM are found by minimizing the multivariate Gaussian log-likelihood for the $i-$th trajectory
\begin{eqnarray}
  L^i(\mathbf{W}_{x},\mathbf{W}_{a},\mathbf{W}_H,\mathbf{W}_{\text{LSTM}},\mathbf{W}_{o}) &=&\nonumber\\  -\sum_{T_{obs}+1}^{T_{pred}}log\left( P([\mathbf{x}^{(i)}_t, \mathbf{a}^{(i)}_t],\mathbf{\mu}^{(i)}_t,\mathbf{\Sigma}^{(i)}_t)\right),
  \label{eq:LLog}
\end{eqnarray}
where $T_{obs}$ is the time frame until when the ground truth data is observed by the LSTM, while $T_{obs}+1,\ldots,T_{pred}$ are the time frames for which is requested the prediction. The loss of Eq.~\ref{eq:LLog} is minimized over all the training sequences, and to prevent overfitting we include an $l_2$ regularization term.

%
%
\subsection{MX-LSTM optimization} \label{sec:opt}
\vspace{-0.1cm}
The optimization provides the weight matrices of the MX-LSTM, which in turn produce the set of Gaussian parameters, including the full covariance $\mathbf{\Sigma}$. The latter is needed to enforce the LSTM in encoding  the relations among the $(x,y)$ coordinate distributions of the tracklets and the vislets, which we will further discuss in Sec.~\ref{sec:dataset}. 

In general, the estimation of a full covariance matrix through optimization of an objective function (as the log-likelihood of Eq.(\ref{eq:LLog}) ) is a difficult numerical problem~\cite{pinheiro1996unconstrained}, since one must guarantee that the resulting estimate is a proper covariance, \emph{i.e.,} a positive semi-definite (p.s.d.) matrix. 

LSTMs involving log-likelihood losses over Gaussian distributions have been restricted so far to two dimensions for simple Gaussian~\cite{alahi2016cvpr} or mixture of Gaussian~\cite{graves2013generating} distributions, in which the $2\times2$ covariance matrices have been obtained by simply optimizing the scalar correlation index $\rho_{x,y}$, which becomes the covariance term of $\mathbf{\Sigma}$ with $\sigma_{x,y}=\rho_{x,y}\sigma_x\sigma_y$ ~\cite{graves2013generating}.
In the case of higher dimensional problems, pairwise correlation terms cannot be optimized and used to build $\mathbf{\Sigma}$, since the optimization process for each correlation term is independent from each other, while the positive-definiteness is a simultaneous constraint on multiple variables~\cite{pourahmadi2011covariance}. 
This lacks of coordination provides matrices far from being s.d.p., that in turns require a correction procedures by projecting into the closest s.d.p. matrix using, for instance, a cost function based on the Frobenious norm~\cite{boyd2005least,higham1988computing}. These procedures are costly~\cite{pinheiro1996unconstrained}, and difficult to be embedded into the optimization process~\cite{dennis1996numerical}, especially in the case of the LSTM, where nonlinearities due to the embedding weights make the analytical derivation hard to formulate. So far, no LSTM loss has involved full covariances of dimension $>2$.

Our solution involves unconstrained optimization, where an opportune parameterization of the variables to learn enforces the positive semi-definite constraint, which is easier to express, dramatically improving the convergence properties of the optimization algorithm.

In practice, we consider the Choleski family of parameterizations~\cite{pourahmadi2011covariance}:
let $\mathbf{\Sigma}$ denote a definite positive $n\times n$ (in our case, $n=4$) covariance matrix. Since $\mathbf{\Sigma}$ is symmetric, only $n(n+1)/2$ parameters are requested to represent it. The Choleski factorization is given by:
\begin{equation}
    \mathbf{\Sigma}=\mathbf{L}^T\mathbf{L},\label{Eq:Chole}
\end{equation}
where $\mathbf{L}$ is a $n\times n$ upper triangular matrix.  In practice, the optimization process would focus on finding the  $n(n+1)/2$ distinct scalar values for $\mathbf{L}$, which then solve for the covariance given Eq. (\ref{Eq:Chole}). One problem with the Cholesky factorization is its non-uniqueness: any matrix obtained by multiplying a subset of
the rows of $\mathbf{L}$ by -1 is valid; as a consequence, non-uniqueness of the solution makes the optimization process hard to converge. 
To make $\mathbf{L}$ unique, its diagonal elements have to be all positive. To this end, the Log-Cholesky parameterization~\cite{pourahmadi2011covariance}  assumes that the values found by the optimizer of the main covariance diagonal are the log of the values of $\mathbf{L}$: Formally,  
the values found by the optimizer can be written as
\[
\mathbf{\theta}_{L}= \begin{bmatrix}
    \log l_{1,1}       & l_{1,2} & l_{1,3} & l_{1,4} \\
     0      & \log l_{2,2} & l_{2,3} &   l_{2,4} \\
    0& 0& \log l_{3,3}  & l_{3,4}\\
    0& 0&0 & \log l_{4,4} 
\end{bmatrix}
\].
In practice, after the estimation of  $\mathbf{W}_{x}$, $\mathbf{W}_{a}$, $\mathbf{W}_H$, $\mathbf{W}_{\text{LSTM}}$, $\mathbf{W}_{o}$ parameters, the values of $\mathbf{\theta}_{L}$ are extracted by  
\begin{equation}
    [\mathbf{\mu}^{(i)}_t,\hat{\mathbf{\theta}_{L}}^{(i)}_t] = \mathbf{W}_o \mathbf{h}^{(i)}_{t-1}, \label{eq:params}
\end{equation}
where $\hat{\mathbf{\theta}_{L}}$ is the vectorized version of $\mathbf{\theta}_{L}$. Then, the diagonal values of $\mathbf{\theta}_{L}$ are exponentiated to form $\mathbf{L}$ and obtaining $\mathbf{\Sigma}$ through Eq.~\eqref{Eq:Chole}. 
\section{Motivation for the MX-LSTM} \label{sec:dataset}
\vspace{-0.1cm}
So far, no quantitative studies focused on how  head pose knowledge impacts on the trajectory forecasting. 
Here, we show a preliminary analysis of the common forecasting datasets with emphasis on the head pose, that motivated the design of the MX-LSTM. 

    \begin{figure*}[t!] 
	\begin{center}
		\includegraphics[width=1\linewidth]{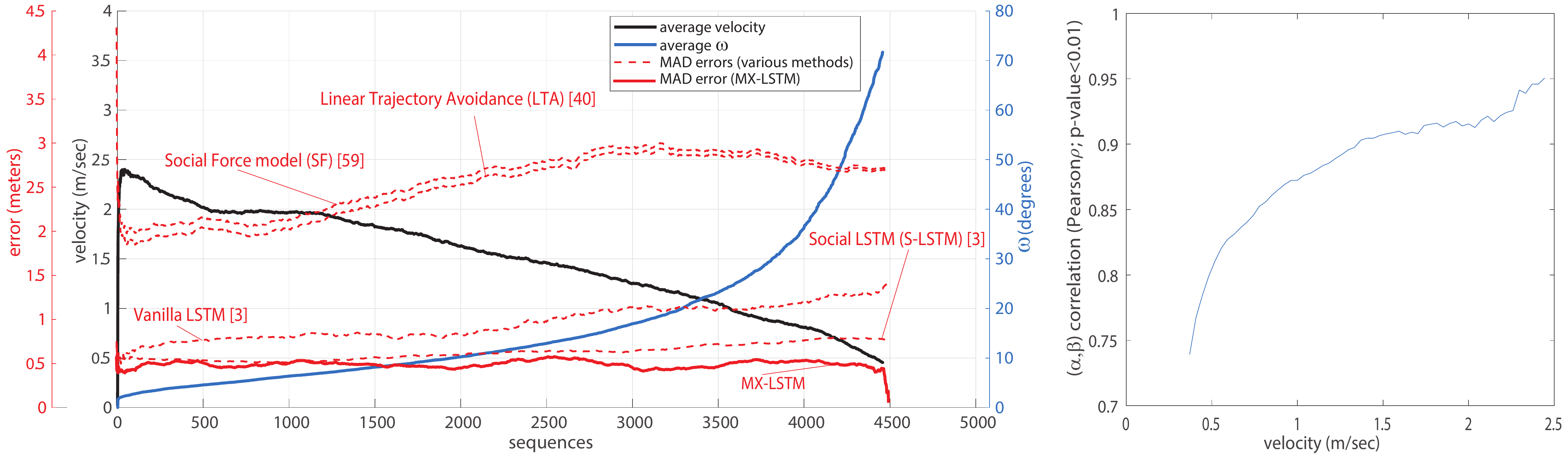}
		\caption{Motivating the MX-LSTM: a) analysis between the angle discrepancy $\omega$ between head pose and movement, the pedestrain smoothed velocity and the average errors of different approaches on the UCY sequence~\cite{lerner2007crowds}; b) correlation between movement angle $\beta$ and head orientation angle $\alpha$ when the velocity is varying (better in color).}\label{fig:angles}
		\vspace{-10pt}
	\end{center}
    \end{figure*}

In particular, we focus on the UCY dataset~\cite{lerner2007crowds}, composed by the Zara01, Zara02, and UCY sequences, which provides the annotations for  the pan angle of the head pose of all the pedestrians.  We also consider the Town Center dataset~\cite{benfold2011stable}, where we manually annotated the head pose, using the same annotation protocol of~\cite{lerner2007crowds}.
We discover the following facts\footnote{Here is presented the analysis on the UCY sequence, which is similar to what we observed on the other sequences.}:
    
    \vspace{.5em}
    \noindent\textbf{1) People often do not watch their steps}. To show this fact, for each individual trajectory composed by $T$ frames (omitting the individual indices), we calculate all the $\alpha_t$, $\beta_t$ and $\omega_t$ of Fig.\ref{fig:explanations}c. The $\alpha_t$ is the head pose pan angle with respect to a given reference system; similarly, $\beta_t$ is the angle of motion, and  $\omega_t$ is showing the discrepancy between the two. 
    For each individual trajectory, we compute the average $\omega = 1/N\sum_{t=1...T}\omega_t$.
    On the multi-y-axis Fig.~\ref{fig:angles}a, we show the $\omega$ value (in degrees) of all the sequences, in an increasing order (blue line and axis). From the figure, we omit those sequences where the speed is below 0.45m/sec.: in those cases the individual is essentially still and the movement vector $\vv{\bm{x_{t+1}-x_{t}}}$ carry few if no meaning, and consequently the angle $\beta_t$ cannot be taken into account. The $\omega$ value ranges from $0.02^{\circ}$ to $72^{\circ}$. We conclude that in ~25\% of the video sequences the misalignment between the head pose and the step direction is larger than
 $20^{\circ}$.
    
    \vspace{.5em}
    \noindent\textbf{2) Head pose and movements are (statistically) correlated};
    On the same figure, we report the velocity curve (black solid line and axis), where each y-point  gives the average speed of the $i-th$ ordered trajectory on the x-axis. For the sake of readability, the curve has been smoothed with moving average filter of size 10.  As it shows, there is a relation of inverse proportionality between the $\omega$ and the pedestrian speed: the alignment between the head towards the direction of movement is higher when the speed is higher; when the person slows down the head pose is dramatically misaligned. The relation is statistically significant: we consider the Pearson circular correlation coefficient~\cite{jammalamadaka2001topics} between the angles $\alpha_{t}$ and $\beta_t$, computed over all the frames of the sequences considered for that figure. On the whole data, the correlation is 0.83 (p-value$<0.01$).
    We also investigate how the correlation changes with the speed: Fig.~\ref{fig:angles}b shows the correlation values against velocity, computed by pooling the $\alpha_{t}$ and $\beta_t$ angles around a certain velocity value; in particular, each correlation value at velocity $\tau$ has been computed by considering all of the samples in the range $[\tau-0.01R, \tau +0.01R]$, where $R$ is the whole velocity range. All the reported values have statistical significance (p-value$<0.01$). The plot shows clearly that the correlation is lower at low velocities, where the discrepancy between the $\alpha_t$ and $\beta_t$ angles is in general higher.  
    The challenge here is to investigate whether this discrepancy can be learned by the MX-LSTM to improve the forecasting. More intriguingly, MX-LSTM should learn how these relations evolve in time, which has not been investigated yet, since the analysis done so far consider each time instant as independent from each other.  
    
    \vspace{.5em}
    \noindent\textbf{3) Forecasting errors are in general higher when the speed of the pedestrian is lower}; In Fig.~\ref{fig:angles} are reported the Mean Average Displacement (MAD) error~\cite{pellegrini2009iccv} (red line and axis) of the following approaches: SF~\cite{yamaguchi2011cvpr}, LTA~\cite{trautman2010iros}, vanilla LSTM and Social LSTM~\cite{alahi2016cvpr}, together with our MX-LSTM approach.  In general, lower velocities bring to higher errors, since when people are walking very slowly their behavior become less predictable, due to physical reasons (less inertia) but also behavioral (people walking slowly are usually involved into other activities, like talking with other people, looking around). On the contrary, it is shown here that  MX-LSTM is performing well  even at lower velocities, reaching errors very close to zero with static people (more details in Sec.\ref{sec:experiments}).

Summarizing, head pose is correlated with the movement, especially when people move fast. When people move slow, the correlation is weaker but significant, the prediction errors are larger, and the head pose is drastically misaligned with the movement. These facts justify and motivate our objective with the MX-LSTM, to capture the head pose information jointly with the movement and use it for a better forecasting.
%
%

\vspace{-0.1cm} \section{Experiments}\label{sec:experiments}
\vspace{-0.1cm}
We present here both quantitative and qualitative experiments.
Quantitative results validate the proposed MX-LSTM model, setting the new state-of-the-art for trajectory forecasting; results are also provided for an ablation study showing the importance of the different parts of the MX-LSTM. Finally, we present the very first results on head pose forecasting. Qualitative results unveil the interplay between tracklets and vislets that the MX-LSTM has learnt.

\begin{table*}[t]
  \centering
  \caption{Mean and Final Average Displacement errors (in meters) for all the methods on all the datasets. The first 5 columns are the comparative methods and our proposed model trained and tested with GT annotations. MX-LSTM-HPE  is our model 
  tested with the output of a real head pose estimator~\cite{lee2015fast}. The last 3 columns are variations of our approach trained and tested on GT annotations.}
  \label{tab:res-oracle}
  \resizebox{\textwidth}{!}{%
  \begin{tabular}{|c|l *{5}{|C{.093\textwidth}} !{\vrule width 2pt} C{.093\textwidth} !{\vrule width 2pt} *{3}{C{.093\textwidth}|}}
    \hline
    Metric & Dataset    &  SF~\cite{yamaguchi2011cvpr}  &  LTA~\cite{pellegrini2009iccv}   &  Vanilla LSTM~\cite{alahi2016cvpr}  & Social LSTM~\cite{alahi2016cvpr} & MX-LSTM  & {\it MX-LSTM-HPE } & Individual MX-LSTM  &  NoFrustum MX-LSTM  &  BD- MX-LSTM   \\
    \hline \hline
    \multirow{5}{*}{\centering MAD} & Zara01 & 2.88 & 2.74 & 0.90 & 0.68 & {\bf 0.59} & {\bf\emph{0.66}} & 0.63 & 0.63 & 0.60 \\
    \cline{2-11}
    & Zara02     & 2.32 & 2.23 & 1.09 & 0.63 & {\bf 0.35} & {\bf\emph{0.37}} & 0.72 & 0.36 & 0.41 \\
    \cline{2-11}
    & UCY        & 2.57 & 2.49 & 0.67 & 0.62 & {\bf 0.49} & {\bf\emph{0.55}} & 0.53 & 0.51 & 0.54 \\
    \cline{2-11}
    & TownCentre &9.35 &9.14   & 4.62 & 1.96 & {\bf 1.15} & {\bf\emph{1.21}} & 2.09 & 1.70 & 1.40 \\
    \hline \hline
    \multirow{5}{*}{\centering FAD} & Zara01 & 5.55 & 5.55 & 1.85 & 1.53 & {\bf 1.31} & {\bf\emph{1.43}} & 1.37 & 1.40 & 1.51 \\
    \cline{2-11}
    & Zara02     & 4.35 &  4.35 & 2.15 & 1.43 & {\bf 0.79} & {\bf\emph{0.82}} & 1.56 & 0.84 & 1.00 \\
    \cline{2-11}
    & UCY        & 4.62 &  4.66 & 1.39 & 1.40 & {\bf 1.12} & {\bf\emph{1.20}} & 1.16 & 1.15 & 1.23 \\
    \cline{2-11}
    & TownCentre & 16.01 & 16.08 & 8.26 & 3.96 & {\bf 2.30} & {\bf\emph{2.38}} & 4.00 & 3.40 & 2.90 \\
    \hline
  \end{tabular}
  }
\end{table*}

\subsection{Quantitative results}\label{sec:quantitative}\vspace{-0.1cm}
We evaluate our model against all the published approaches which made their code publicly available: 
Social Force model (SF)~\cite{yamaguchi2011cvpr}, Linear Trajectory Avoidance (LTA)~\cite{pellegrini2009iccv}, Vanilla LSTM and Social LSTM (S-LSTM)~\cite{alahi2016cvpr}.

Experiments follow the widely-used evaluation protocol of~\cite{pellegrini2009iccv}, in which the algorithm first observes 8 ``observation'' ground truth (GT) frames of a trajectory, predicting the following 12 ones. 
For the three UCY sequences three models have been trained: for each one we used two sequences as training data and then we tested on the third sequence. For Town Centre dataset the model has been trained and tested on the respective provided sets. The grid for the social pooling (Eq.~\eqref{eq:social}) has $N_o\times N_o$ cells with $N_o=32$.
The view frustum aperture angle has been cross-validated on the training partition of the TownCenter and kept fixed for the remaining trials ($\gamma=40^{\circ}$), while the depth $d$ is simply bounded by the social pooling grid. 
Trajectory prediction performances are analyzed with the Mean Average Displacement (MAD) error (euclidean distance between predicted and GT points, averaged over the sequence), and Final Average Displacement (FAD) error (distance between the last predicted point and the corresponding GT point)~\cite{pellegrini2009iccv}.

Results are reported in Table~\ref{tab:res-oracle}.
The MX-LSTM outperforms the state-of-the-art methods in every single sequence and with both  metrics, with an average improvement of 32.7\%. The highest relative gain is achieved in Zara02 dataset, where complex non-linear paths are mostly caused by standing conversational groups and people that walk close to them, avoiding collisions. People slowing down and looking at the window shops pose also a challenge. As shown in Fig.~\ref{fig:angles}, slow moving and interacting pedestrians cause troubles to the competing methods, while MX-LSTM clearly overcomes such shortcomings denoting a better model. Qualitative motivations will follow in Sec.~\ref{sec:qualitative}.

Please note that different methods rely on different input data: 
both SF and LTA require the destination point of each individual, while SF also requires annotations about social groups; MX-LSTM requires the head pose of each individual for the first 8 frames, but this can be estimated by a head pose estimator. This motivates our next experiment: we automatically estimate the head bounding box given the feet positions on the floor plane, assuming  people being $1.80$m tall. Then, we apply the head pose estimator of~\cite{lee2015fast} which gives continuous angles that can be used as input of our approach now named ``MX-LSTM-HPE''.   
As shown by the scores in Table~\ref{tab:res-oracle}, MX-LSTM-HPE does not suffer about small errors in the input head pose, with an average drop in performances of only 5\%. Note that MX-LSTM-HPE still outperforms all competing methods on all dataset even with the noisy estimated head pose information.

How accurate should the head pose estimation be, for the MX-LSTM-HPE to have convincing performances, for example outperforming the Social LSTM? We answer this question by corrupting the true head pose estimate with additive Gaussian noise$\sim\mathcal{N}(\alpha_t,\hat{\sigma})$, where $\alpha_t$ is the correct head pose and $\hat{\sigma}$ the standard deviation. MX-LSTM-HPE outperforms social-LSTM up to a noise of $\hat{\sigma}=24^{\circ}$.

\vspace{-0.2cm}
\subsubsection{Ablation study}\label{sec:ablation}
Aside with the models in the literature, we investigate three variations of the MX-LSTM to capture the net contributions of the different parts that characterize our approach.

\noindent\textbf{Block-Diagonal MX-LSTM (BD-MX-LSTM)}: it serves to highlight the importance of estimating full covariances to understand the interplay between tracklets and vislets.  
%
%
Essentially, the approach estimates two bidimensional covariances\footnote{The $2\times2$ covariance is estimated employing two variances $\sigma_1, \sigma_2$ and a correlation terms $\rho$ as presented in \cite{graves2013generating} Eq.(24) and (25).} $\mathbf{\Sigma}_x$ and $\mathbf{\Sigma}_a$ for the trajectory and the vislet modeling respectively, without capturing the cross-stream covariances.

\noindent\textbf{NoFrustum MX-LSTM}: this variation of the MX-LSTM uses social pooling as in~\cite{alahi2016cvpr}, in which the interest area where people hidden states $\{\mathbf{h}^{j}_t\}$ are pooled into the social tensor  all around the individual. In other words, no frustum selecting the people that have to be considered is used here.

\noindent\textbf{Individual MX-LSTM}: In this case, no social pooling is taken into account, therefore the embedding operation of Eq.~(\ref{eq:pooling}) is absent, and the weight matrix $\mathbf{W}_H$ vanishes. In practice, this variant learns independent models for each person, each one considering the tracklet and vislet points.

\vspace{1em}
Table~\ref{tab:res-oracle}, last three columns, reports numerical results for all the MX-LSTM simplifications on all the datasets. The main facts that emerge are: 1) the highest variations are with the Zara02 sequence, where MX-LSTM doubles the performances of the worst approach (Individual MX-LSTM); 2) the worst performing is in general Individual MX-LSTM, showing that social reasoning is indeed necessary; 3) social reasoning is systematically improved with the help of the vislet-based view-frustum; 4) full covariance estimation has a role in pushing down the error which is already small with the adoption of vislets.  

Summarizing the results so far, having vislets as input allows to definitely increase the trajectory forecasting performance, even if vislets are estimated with noise. Vislets should be used to understand social interactions with social pooling, by building a view frustum that tells which are the people currently observed by each individual. All of these features are done efficiently by the MX-LSTM: in fact the training time is the same with having an LSTM with social pooling.

\vspace{-0.2cm}
\subsubsection{Head pose forecasting}\vspace{-0.1cm}
As done with trajectories, we are also providing a forecast of the head pose of each individual at each frame which is a distinctive attribute of our method.   We evaluate the performances of this estimation in terms of mean angular error $e_{\alpha}$, which is the mean absolute difference between the estimated pose (angle $\alpha_{t,\cdot}$ in Fig.~\ref{fig:explanations}c) and the annotated GT.


\begin{table}[t!]
  \centering
  \small
  \caption{Mean angular error (in degrees) for the state-of-the-art head pose estimator~\cite{lee2015fast}, and the MX-LSTM model fed with GT annotations and estimated values (MX-LSTM-HPE).}
  \label{tab:res-angle}
  \resizebox{\linewidth}{!}{
  \begin{tabular}{|l|*{4}{C{.06\textwidth}|}}
    \hline
    Metric                &  Zara01  &  Zara02  &   UCY   &  Town Centre \\
    \hline
    HPE~\cite{lee2015fast} & 14.29   &   20.02  &  19.90 &   25.08    \\   
    \hline
    MX-LSTM               &  12.98   &   20.55  &  21.36  &   26.48    \\
    \hline
    MX-LSTM-HPE  &  17.69   &   21.92  &  24.37  &   28.55  \\
    \hline
  \end{tabular}
  }
\end{table}


Table~\ref{tab:res-angle} shows numerical results of the static head pose estimator~\cite{lee2015fast} (HPE), the MX-LSTM using GT head poses, and the MX-LSTM fed with the output of HPE during the observation period (MX-LSTM-HPE).
In all the cases our forecast output is comparable with the one of HPE, but in our case we do not use appearance cues -- \ie we do not look at the images at all.
In case of Zara01, the MX-LSTM is even better that the static prediction showing the forecasting power of our model. In our opinion this is due to the fact that in this sequence trajectories are mostly very linear and fast, and heads are mostly aligned with the direction of motion. When we provide estimations to the MX-LSTM model during the observation period, angular error increases, as expected. Despite this, the error is surprisingly limited.

\subsection{Qualitative results}\label{sec:qualitative}
\vspace{-0.1cm}
    \begin{figure*}[t!] 
	\begin{center}
		\includegraphics[width=1\linewidth]{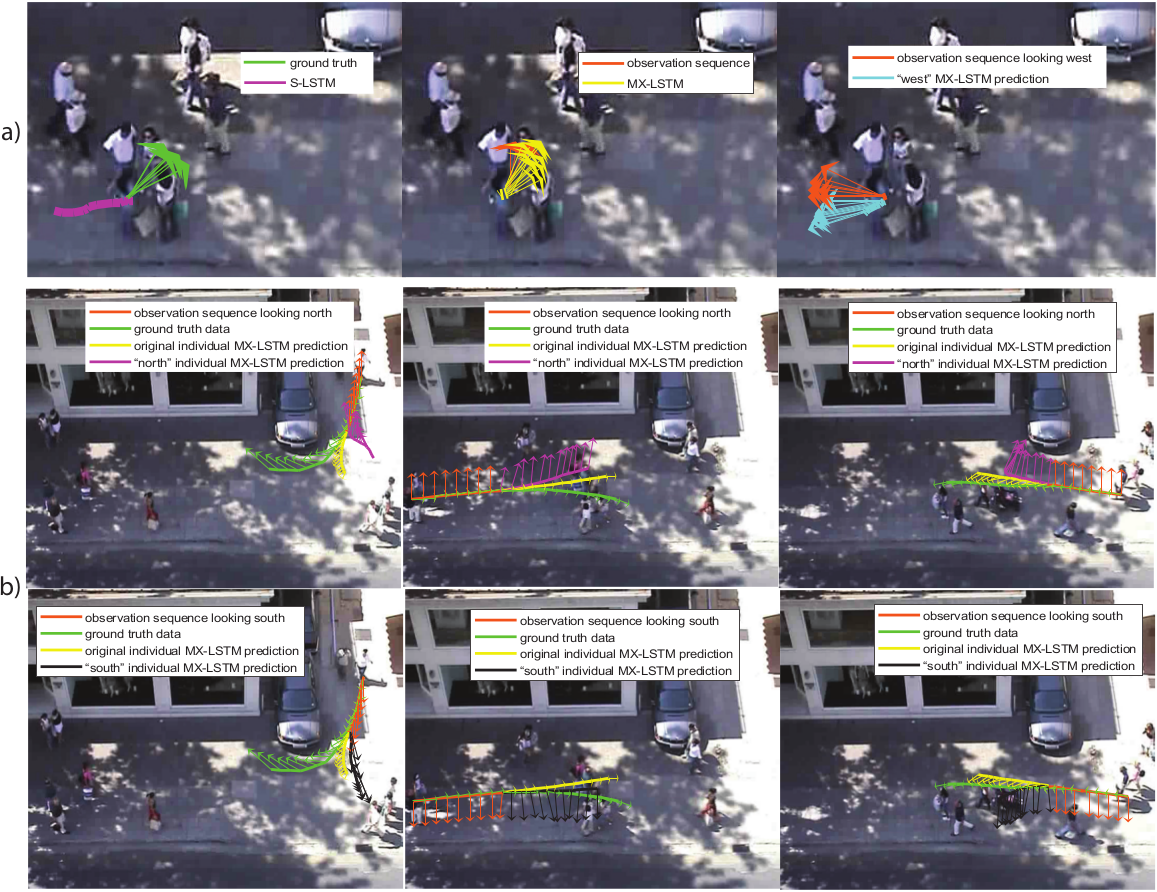}
		\caption{Qualitative results: a) MX-LSTM  b) Ablation qualitative study on Individual MX-LSTM (better in color).}\label{fig:Qualitative}
		\vspace{-15pt}
	\end{center}
    \end{figure*}

Fig.~\ref{fig:Qualitative} shows  qualitative results on the Zara02 dataset, which has been shown to be the most complex scenario in the quantitative experiments
%
Fig.~\ref{fig:Qualitative}a presents MX-LSTM results: a group scenario is taken into account, with the attention focused on the girl in the bottom-left corner. In the left column, the green ground-truth prediction vislets show that the girl is conversing with the group members, with a movement close to zero and the pan head angle which oscillates. In magenta, the behavior of the S-LSTM, predicting erroneously the girl leaving the group. This error confirms the problem of competing methods in forecasting the motion of people slowly moving or static as discussed in Sec.\ref{sec:dataset}, and further confirmed by the results of the quantitative experiments. 
In the central column, the observation sequence given to the MX-LSTM is shown in orange (almost static with oscillating vislets). The related prediction (in yellow) shows oscillating vislets, and almost no movement, confirming that the MX-LSTM has learnt this particular social behavior. If we provide to MX-LSTM an artificial observation sequence with the annotated positions (real trajectory) but vislets oriented toward west  (third column, orange arrows), where no people are present, the MX-LSTM predicts a trajectory departing from the group (cyan trajectory and arrows).

The two rows of Fig.~\ref{fig:Qualitative}b) analyze the Individual MX-LSTM, in which no social pooling is taken into account. Therefore, here each pedestrian is not influenced by the surrounding people, and the relationship between the tracklets and the vislets in the prediction can be observed without any confounding factor. 
Fig.~\ref{fig:Qualitative}b) first row shows three situations in which the vislets of the observation sequence are artificially made pointing north (orange arrows), resulting not aligned with the trajectory. In this case the Individual MX-LSTM predicts a decelerating trajectory drifting toward north (magenta trajectory and vislets), especially in the second and third pictures.  If the observation has the legit vislets (green arrows, barely visible since they are aligned with the trajectory), the resulting trajectory (yellow trajectory and vislets) has a different behavior, closer to the GT (green trajectory and vislets). The second row is similar, with the observation vislets made pointing to south. The prediction with the modified vislets is in black. The only difference is in the bottom left picture: here the observation vislets pointing south are in accord with the movement, so that the resulting predicted trajectory is not decelerating as in the other cases, but accelerating toward south.

\section{Conclusion} \label{sec:conc}

This paper showed that sequences of consecutive head poses, \emph{i.e.}, the vislets, are of great help for trajectory forecasting. We introduced a model to incorporate vislets and tracklets, the MX-LSTM, which mixes together the two streams of information providing cross-stream $4 \times 4$ covariances, that explain how head poses and positions on the plane are correlated, providing accurate forecasting prediction for both of them. This has been possible thanks to an optimization process embedded into the LSTM backpropagation which uses a log-Cholesky parameterization, leading to unconstrained optimization.
We believe that consideration of vislets would allow us, in future work, to also encode specific areas of interest into the trajectory forecasting.

\noindent \textbf{Acknowledgements}:
This work has received funding from the European Union's Horizon 2020 research and innovation programme under the Marie Sklodowska-Curie Grant Agreement No. 676455, and has been partially supported by the POR FESR 2014-2020 Work Program (Action 1.1.4, project No.10066183).

{\small
\bibliographystyle{ieee}
\bibliography{egbib}
}

\end{document}